\documentclass[conference]{IEEEtran}

\usepackage{float}
\usepackage[pdftex]{graphicx}
\usepackage{amsmath,amssymb} 
\usepackage{bm}
\usepackage{subfigure}
\usepackage{cite}

\ifCLASSINFOpdf
\else
\fi
\begin{document}

\pagestyle{headings}

\def\ICSPACSubNumber{116}  

\title{Deep Learning Representation using Autoencoder for 3D Shape Retrieval} 

%





\author{\IEEEauthorblockN{Zhuotun Zhu, Xinggang Wang, Song Bai, Cong Yao, Xiang Bai}\\
Department of Electronics and Information Engineering\\
Huazhong University of Science and Technology, PR China\\
{\{zhuzhuotun, xgwang, songbai}\}@hust.edu.cn, yaocong2010@gmail.com, xbai@hust.edu.cn
}


%


\maketitle

\begin{abstract}
We study the problem of how to build a deep learning representation for 3D shape.
Deep learning has shown to be very effective in variety of visual applications, such as image classification and object detection.
However, it has not been successfully applied to 3D shape recognition.
This is because 3D shape has complex structure in 3D space and there are limited number of 3D shapes for feature learning.
To address these problems, we project 3D shapes into 2D space and use autoencoder for feature learning on the 2D images.
High accuracy 3D shape retrieval performance is obtained by aggregating the features learned on 2D images.
In addition, we show the proposed deep learning feature is complementary to conventional local image descriptors.
By combing the global deep learning representation and the local descriptor representation, our method can obtain the state-of-the-art performance on 3D shape retrieval benchmarks.
\end{abstract}


%
\IEEEpeerreviewmaketitle

\section{Introduction}

With the fast development of 3D printer, Microsoft Kinect sensor and laser scanner, etc., there are more and more digitized 3D models that need to be recognized.
Thus it is critical to study how to build an efficient 3D shape search engine.
However, due to the  intrinsic complex structure of 3D shape, it is hard to handle 3D shape using a simple representation for efficient search.

Along with the development of computer vision and machine learning,
deep learning methods have been proven to be very effective for
visual recognition. For example, deep convolutional neural network
(CNN) \cite{lecun1989backpropagation} has achieved the
state-of-the-art performance for object recognition on the ImageNet
dataset \cite{krizhevsky2012imagenet} and for object detection on
the PASCAL dataset \cite{girshick2013rich}. One reason of the
success of deep learning for visual recognition is that the deep
learning methods can automatically learn the features with the
superior discriminatory power for image representation, rather than
using hand-crafted image descriptors. Currently, in the context of
3D shape recognition, shape descriptors are mainly hand-crafted and
deep learning representation has not been widely applied. It seems
that it is hard to directly apply deep learning methods to 3D shape
representation, since deep learning methods need a large amount of
data to bridge the visual gap among training examples from the same
object category; and it is unlikely to learn a good representation
using a few data with large visual variation.

The above developments of deep learning are in a supervised way and
are not suitable for retrieval task. From the aspect of unsupervised
deep learning, Hinton and Krizhevsky~\cite{Krizhevsky} proposed the
autoencoder algorithm with the application of image retrieval, which
is then used for some other specific tasks like face alignment
\cite{Face Alignment}.
Training autoencoder does not require any label information. The
autoencoder can be regarded as a multi-layer sparse coding network.
Each node in the autoencoder network can be regarded as a prototype
of object image/shape. From the bottom layer to the top layer, the
prototype contains richer semantic information and becomes a better
representation. After the autoencoder network is learnt, the
coefficients obtained by reconstructing image/shape based on
prototypes are used as feature for 3D shape matching and retrieval.
Since the autoencoder can learn feature adaptively to training data,
it can get excellent performance for image retrieval.

\begin{figure}[!ht]
  \centering
  \includegraphics[width=0.4\textwidth]{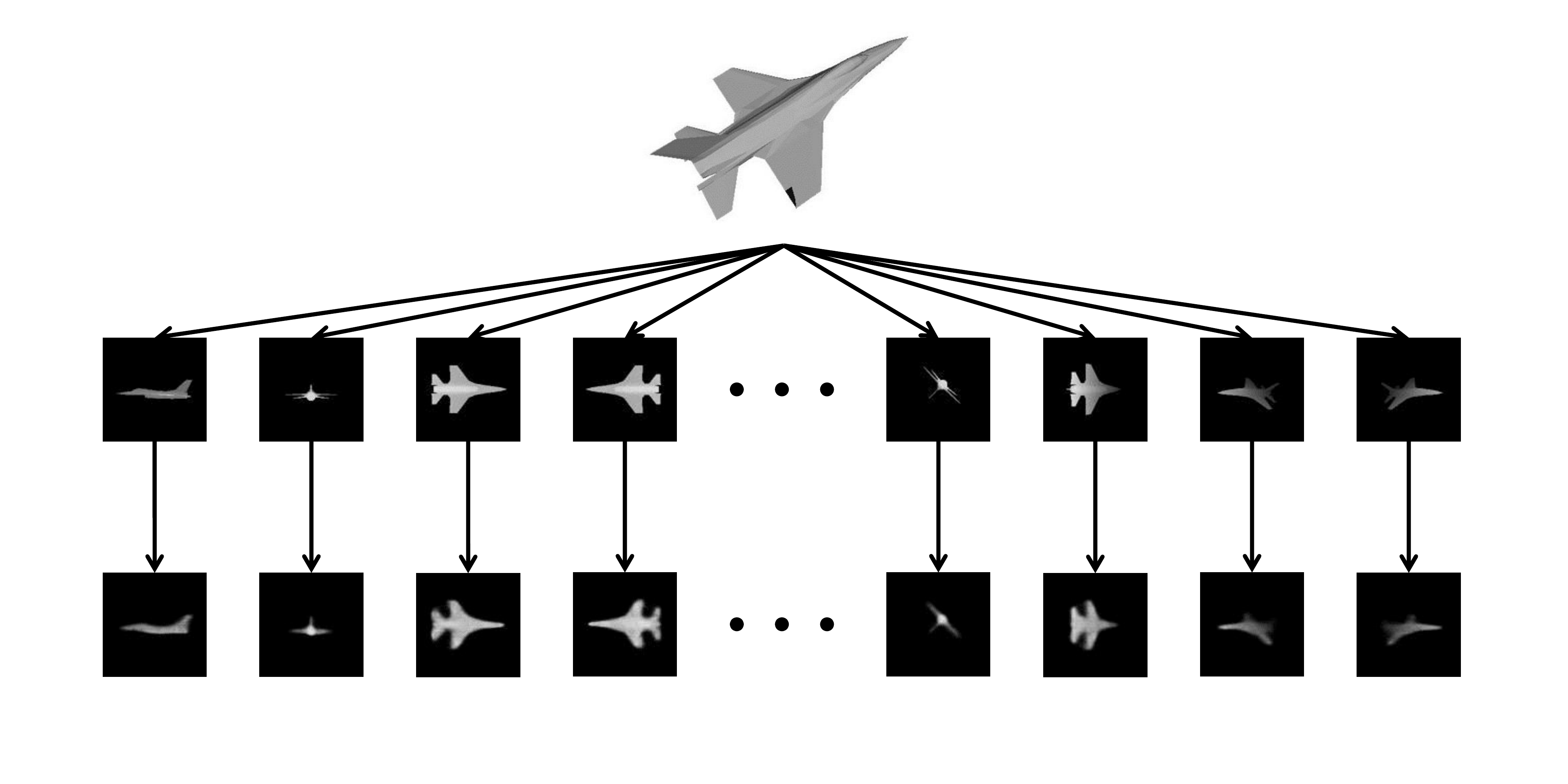}\\
  \caption{A specific illustration of our method to reconstruct 2D images. Note that the first row displays the original depth images in gray-scale of the 3D shape, while the second row shows the reconstructed ones corresponding to the images of the first row. And the black dots indicates those extracted from other different views.}    \label{fig:intromodel}
 \end{figure}

Motivated by the view-based 3D shape methods \cite{LFD, CM-BoF},  in
which a 3D shape can be projected into many 2D depth images, we aim
to use autoencoder to learn a 3D shape representation based on the
depth images obtained by projection. As shown in
Fig.~\ref{fig:intromodel}, a 3D shape is projected into many
different depth images; the learnt autoencoder can reconstruct the
depth images nicely. Matching 3D shape based on the autoencoder
features can be converted to a set-to-set matching problem,
conventional set-to-set distance, like the Hausdorff distance, can
be adopted. Our autoencoder based 3D shape representation is a deep
learning representation; compared to the representations based on
local descriptor, e.g. SIFT, it is a global representation. This
global deep learning representation and the representation based on
local descriptors are complementary to each other.

In summary, the main contributions of this paper are: (1) A new method to learn deep learning representation for 3D shape using autoencoder;
(2) combining the global deep learning representation with local descriptor representation and obtaining the state-of-the-art 3D shape retrieval performance.



\section{Related Work}\label{sec:RW}

Based on the main idea that ``two 3D models are similar if they look similar with each other from all viewing angles'', there are plenty of view-based approaches that have been regarded as the most discriminative methods on literature~\cite{princeton}. Since our shape descriptor is also view-based, we mainly discuss some effective, competing view-based approaches during the following part.

In~\cite{SGM}, Cyr and Kimia recognized a 3D shape by comparing a view of the shape with all views of 3D objects using shock graph matching.
Ohbuchi et al.~\cite{BoF} utilized local visual features by using the Scale Invariant Feature Transform~\cite{loweSIFT} to retrieve 3D shapes. A host of local features describing the 3D models is integrated into a histogram using Bag-of-Features~\cite{fei2005bayesian} to reduce the computation complexity.
Vranic~\cite{Desire} presented a composite 3D shape feature vector (DESIRE) which consists of depth buffer images, silhouettes and ray-extents of a polygonal mesh. The composite of various feature vectors extracted in a canonical coordinate frame generally performs better than the single method which relies on pairwise alignment of 3D objects.
Later on, Papadakis et al.~\cite{Hybrid} made use of a hybrid descriptor (Hybrid) which consists of both depth buffer based 2D features and spherical harmonies based 3D features. The Hybrid adopts two alignment methods to compensate inner rotation variance and then uses Huffman coding to further compress feature descriptors.
Also, they presented a 3D descriptor (PANORAMA)~\cite{Panorama} that captures the panoramic view of a 3D shape by projecting it to a lateral surface of a cylinder parallel to one of its three principal axes.
By aligning its principle axes to capture the global information and combining
2D Discrete Fourier Transform and 2D Discrete Wavelet Transform, the
PANORAMA outperforms all the other 3D shape retrieval methods on
several standard 3D benchmarks. Meanwhile, Lian et al.~\cite{CM-BoF}
used Bag-of-Features and Clock Matching (CM-BoF) on a set of
depth-buffer views obtained from the projections of the normalized
object. The CM-BoF method also takes advantage of the preserved
local details as well as isometry-invariant global structure to
reach a competing result. Prior to that, they also proposed a shape
descriptor named Geodesic Sphere based Multi-view Descriptors
(GSMD)~\cite{GSMD} measuring the extend to which a 3D polygon is
rectilinear based on the maximum ratio of the surface area to the
sum of three orthogonal projected areas. Recently, Bai et
al.~\cite{ShapeVocabulary} adopted contour fragments as the input
features for learning a BoW model, which is general and efficient
for both 2D and 3D shape matching.


\section{Deep Learning Representation using Autoencoder}\label{sec:AE}
\begin{figure}[!ht]
  \centering
  \includegraphics[width=1\linewidth]{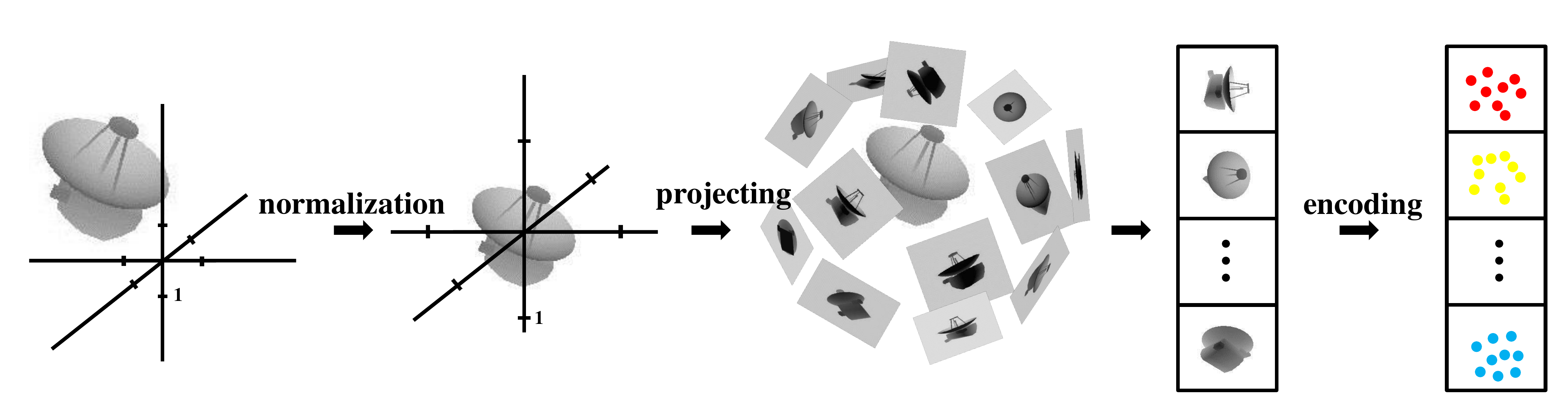}\\
  \caption{The flow chart of 3D shape representation using autoencoder.
  First, we conduct pose normalization for differences in translation and scale to each 3D model. Next, each 3D shape is represented by a set of depth-buffer images. Finally all the projections are used to train the autoencoder to acquire the code as a low-dimensional representation of the depth images, based on which to conduct 3D shape retrieval. In the last image, the colored dots indicate those features extracted from the corresponding depth images.}
  \label{fig:flow}
\end{figure}

In this Section, given a 3D shape model $S$, we show how to perform DBN-initialized autoencoder for $S$ and then conduct 3D shape retrieval based on the calculated shape code. As shown in Fig.~\ref{fig:flow}, we illustrate a specific flow chart about the whole procedure.

\subsection{Depth Projection Image}\label{sec:DepthPro} 
Different from shapes of 2D images, 3D models represent the 3D objects using a collection of points in 3D space, connected by various geometric entities such as lines, curved surfaces, etc. In our method, the autoencoder initialized by a DBN described in Section~\ref{sec:DBN} is used to reconstruct the gray-scale depth 2D images as input and acts as a low-dimensional coding method. Thus, projecting a 3D model to a collection of 2D images is required to make it possible. For a 3D shape model \textit{S} preprocessed by scale and translation normalization, from a host of angles of view, we collect 2D projections set of \textit{S} defined as
\begin{equation}\label{eq:proj}
   \textbf{P}(S)=\lbrace\textit V_1,\textit V_2,\dots,\textit V_{Np}\rbrace,
\end{equation}
where $Np$ denotes the number of projections for each model.

\begin{figure}[!ht]
  \centering
  \includegraphics[width=0.4\linewidth]{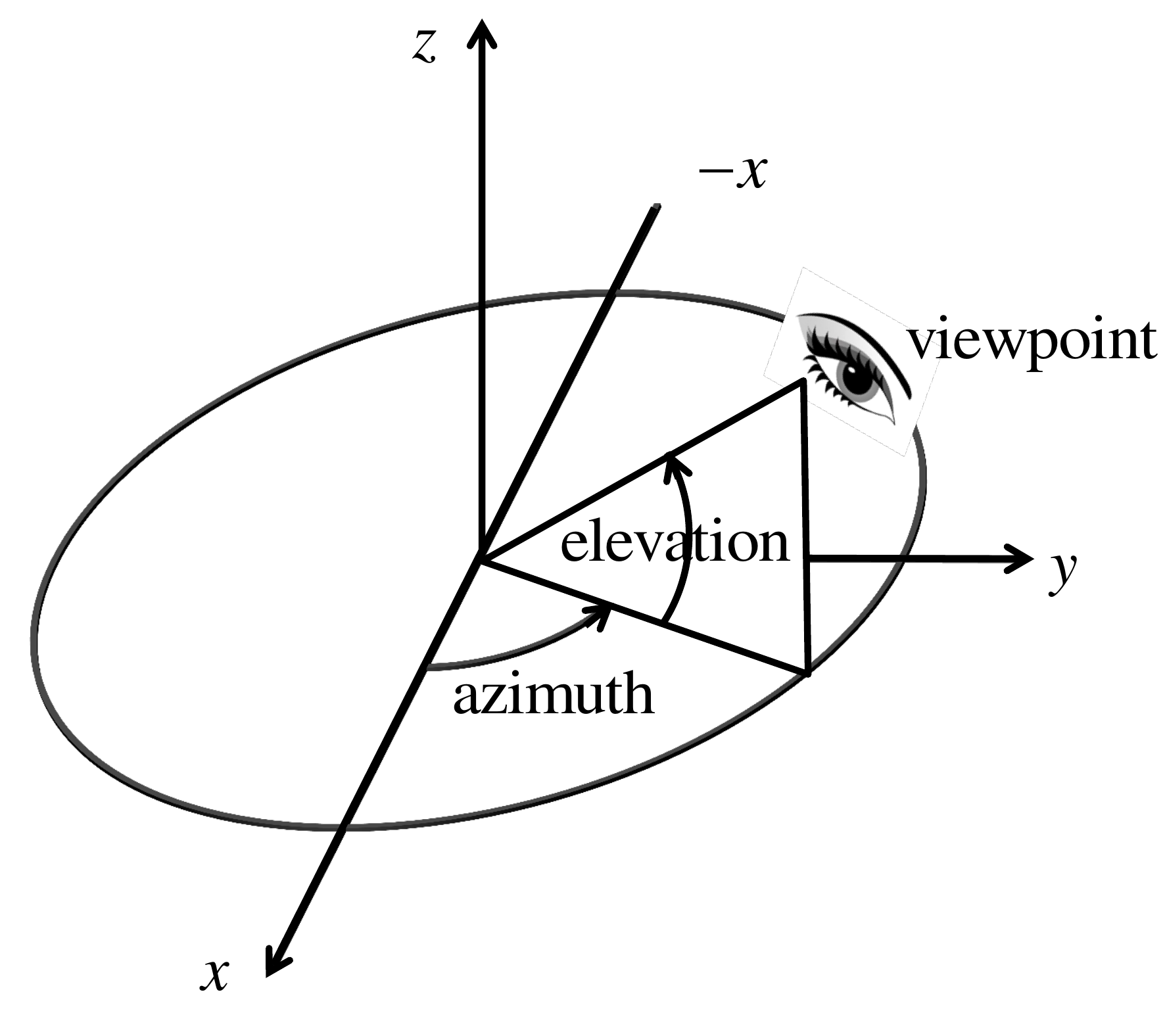}\\
  \caption{The illustration of how we get the projections of a 3D shape model $S$. Azimuth is the polar angle in the \textit{x-y} plane, with positive number indicating anticlockwise rotation of the viewpoint. As for elevation, positive and negative numbers are the angle above and below the \textit{x-y} plane respectively.
  }    \label{fig:projection}
\end{figure}

More specifically, Fig.~\ref{fig:projection} illustrates how we obtain a series of projections for the shape $S$ viewed from different angles both in azimuth and elevation.

\subsection{Deep Belief Network}\label{sec:DBN}
The DBN~\cite{BengioDBN,HintonDBN,LeCunDBN} is a generative graphical model, or alternatively a type of deep neural network, composed of multiple layers of latent variables (``hidden units''), with connections between the layers but not between units within each layer. When trained on plenty of examples in an unsupervised way, a DBN can probabilistically reconstruct the inputs by learning a stack of Restricted Boltzmann Machines (RBMs), where each of the previous RBM's hidden layer serves as the visible layer for the next. That is to say, each time a new RBM is added to the stacked structure of DBN, then the new DBN has a better variational lower bound in the log probability of the data than the previous DBN~\cite{Krizhevsky}.

\begin{figure}[!ht]
  \centering
  \includegraphics[width=0.4\linewidth]{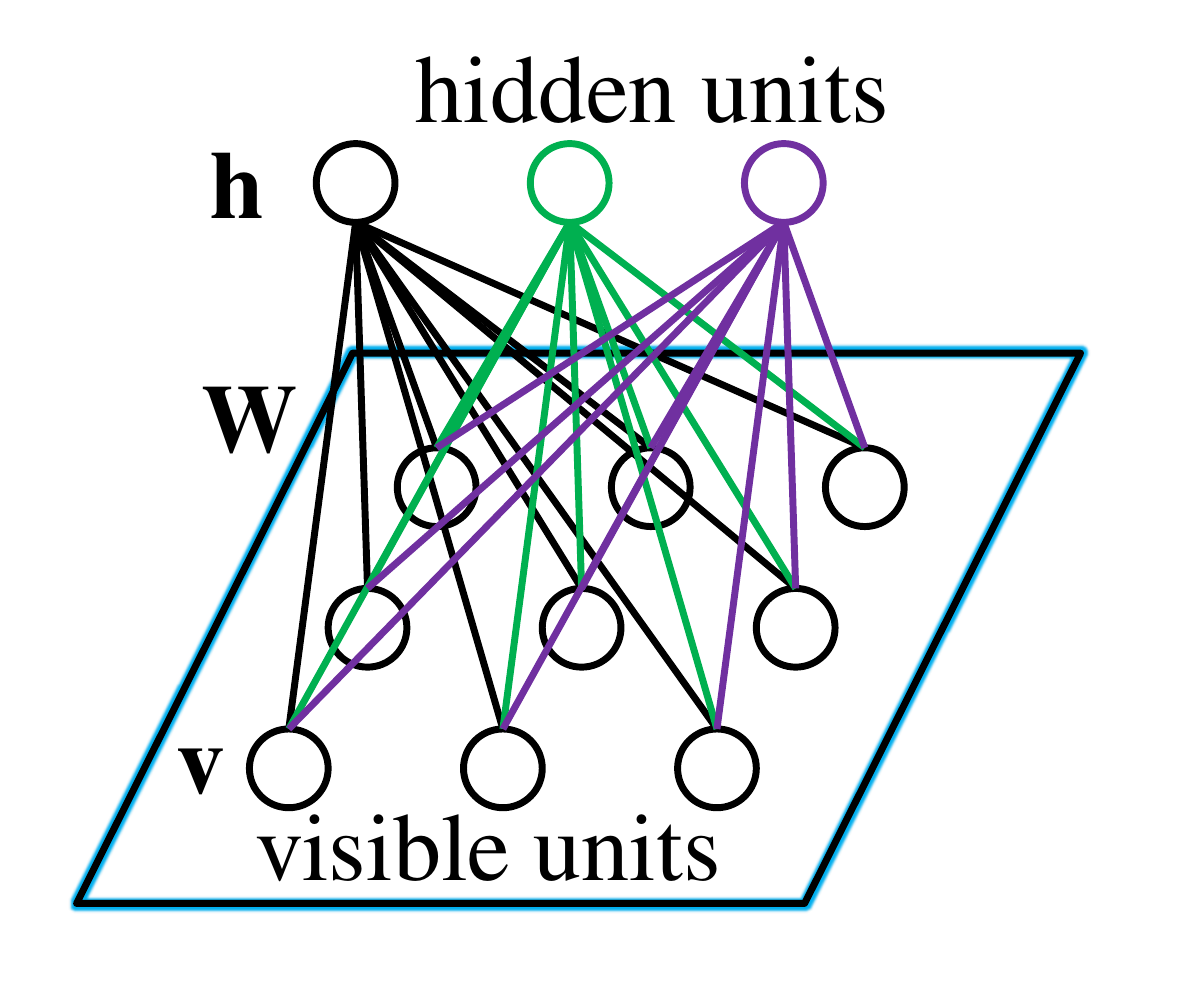}\\
  \caption{A graphical description of RBM. Note that a standard type of RBM has binary-valued visible and hidden units with weights of the connection between them. What needs to be specially emphasized is that there are none connections within visible units or hidden ones, which leads to a property that the hidden unit activations are mutually independent given the activations of visible units and conversely.}    \label{fig:RBM}
\end{figure}

We introduce the ``pretraining'' procedure as shown in Fig.~\ref{fig:RBM} for binary units, then generalize to real-valued units and show that it works well. The pixels correspond to the ``visible units'' since their states can be observed; as for the feature detectors, they correspond to the ``hidden units''. The energy of a joint configuration ($\bm{v}$,$\bm{h}$) for the visible and hidden units is defined in~\cite{Hopfield} as
\begin{equation}\label{eq:Energy}
   \textit{E}(\bm{v},\bm{h})=- \sum_{i\in{visible}}a_iv_i-\sum_{j\in{hidden}}{b_jh_j}-\sum_{i,j}w_{ij}v_ih_j,
\end{equation}
where $v_i$, $h_j$ denote the binary states of visible unit \textit{i} and hidden unit \textit{j} respectively; $a_i$, $b_j$ are their biases and $w_{ij}$ is the connection weight between them.

The network assigns a probability to every possible couple of a visible vector and a hidden one by the following function
\begin{equation}
   \textit{p}(\bm{v},\bm{h})=\frac{1}{Z}e^{-\textit{E}(\bm{v},\bm{h})},
\end{equation}
where the ``partition function'' \textit{Z} is given by the sum of all possible pairs between visible and hidden vectors
\begin{equation}
   \textit{Z}=\sum_{\bm{v},\bm{h}}e^{-\textit{E}(\bm{v},\bm{h})}.
\end{equation}

The probability that the network assigns to a visible vector, is defined as the sum of all possible hidden vectors
\begin{equation}
   \textit{p}(\bm{v})=\frac{1}{Z}\sum_{\bm{h}}e^{-\textit{E}(\bm{v},\bm{h})}.
\end{equation}

The probability of a training image can be increased by adjusting the biases and weights to lower the energy of that image but to increase the energy of the rest, especially for these that own low energy and thus are assigned high probability by the network and make great contribution to the partition function.
The mathematically derived derivative of the log probability of a visible vector to a weight is simple:
\begin{equation}
   \frac{\partial \log p(\bm{v})}{\partial w_{ij}}=\langle{v_ih_j}\rangle_{data}-\langle{v_ih_j}\rangle_{model},
\end{equation}
where the angle brackets denote expectations under the exact distribution specified by the subscript that follows.
Thus, utilizing stochastic steepest ascent as the learning approach is a very simple way in the log probability of training data
\begin{equation}
   \Delta{w_{ij}}=\epsilon(\langle{v_ih_j}\rangle_{data}-\langle{v_ih_j}\rangle_{model}),
\end{equation}
where the $\epsilon$ is the learning rate.

Because of the RBM's restricted structure that there are no direct connections within hidden units, it is pretty easy to obtain an unbiased sample of $\langle{v_ih_j}\rangle_{data}$. Given a training image as the visible vector $\bm{v}$, the binary state ${h_j}$ of every hidden unit $\textit{j}$ is set to 1 with the probability
\begin{equation}
\label{eq:v2h}
   \textit{p}({h_j=1}\mid\bm{v})=S({b_j}+\sum_{i\in{visible}}w_{ij}v_i),
\end{equation}
where $S({x})$ denotes the sigmoid function defined by the formula $1/[1+\exp(-x)]$.

Given a hidden vector $\bm{h}$, it is also quite easy to obtain an unbiased sample of a visible unit's state as a consequence of no connections within visible units. The first equation corresponds with the construction of binary visible units and the second one with linear visible units, where $\textit{N}(\mu,\sigma)$ is a Gaussian with mean value $\mu$ and standard deviation $\sigma$.
\begin{flalign}\label{eq:h2v}
\begin{split}
   &\textit{p}({v_i=1}\mid\bm{h})=S({a_i}+\sum_{j\in{hidden}}w_{ij}h_j), or \\ &v_i=\textit{N}({a_i}+\sum_{j\in{hidden}}w_{ij}h_j,1).\\
\end{split}
\end{flalign}

Obtaining an unbiased sample of $\langle{v_ih_j}\rangle_{model}$, however, is much more tough. It can be done by beginning with any random state of a visible vector and performing alternating Gibbs sampling for quite a long time. One iteration of Gibbs sampling is used to update all the hidden units in parallel applying Eq.~\eqref{eq:v2h} followed by updating all the visible units in parallel applying Eq.~\eqref{eq:h2v}.

Fortunately, a much faster learning algorithm was proposed in \cite{hinton2002training}. This algorithm begins by setting the visible units' states to a training vector. Then the whole hidden units' binary states are calculated in parallel applying Eq.~\eqref{eq:v2h}. After those binary states have been probabilistically chosen for the hidden units, a ``confabulation'' is produced via setting each visible unit ${v_i}$ to 1 with probability as in Eq.~\eqref{eq:h2v}. Update the states of the hidden units once more in order that they can represent features of the confabulation. Then the adjustment of the weight is formulated by
\begin{equation}
   \Delta{w_{ij}}=\epsilon(\langle{v_ih_j}\rangle_{data}-\langle{v_ih_j}\rangle_{recon}),
\end{equation}
where the $\langle{v_ih_j}\rangle_{data}$ is the fraction of times that the visible unit \textit{i} and the hidden unit \textit{j} are on together when the hidden units are driven by data, and $\langle{v_ih_j}\rangle_{recon}$ is the corresponding part given by the confabulation. A same learning rules is used to adjust the biases.

In our experiments, this fast learning procedure works out well even though it is just approximating the derivative of the log probability with respect to the training data. 

\subsection{Fine-tuning the Autoencoder}\label{sec:FineTune}
After pretraining a DBN which acts as initialization of an autoencoder, a global fine-tuning procedure replaces the former stochastic, binary activities with crucial, real-valued probabilities and uses backpropagation through the whole structure of autoencoder to adjust the weights as well as biases for a reconstruction model. By minimizing the root mean squared reconstruction error $\sqrt{\sum_i({\langle{v_i}\rangle}_{data}-{\langle{v_i}\rangle}_{recon})^{2}}$, we finally obtain a deep-structured, optimal reconstruction model of the 2D depth images as input.

\begin{figure}[!ht]
  \centering
  \includegraphics[width=0.8\linewidth]{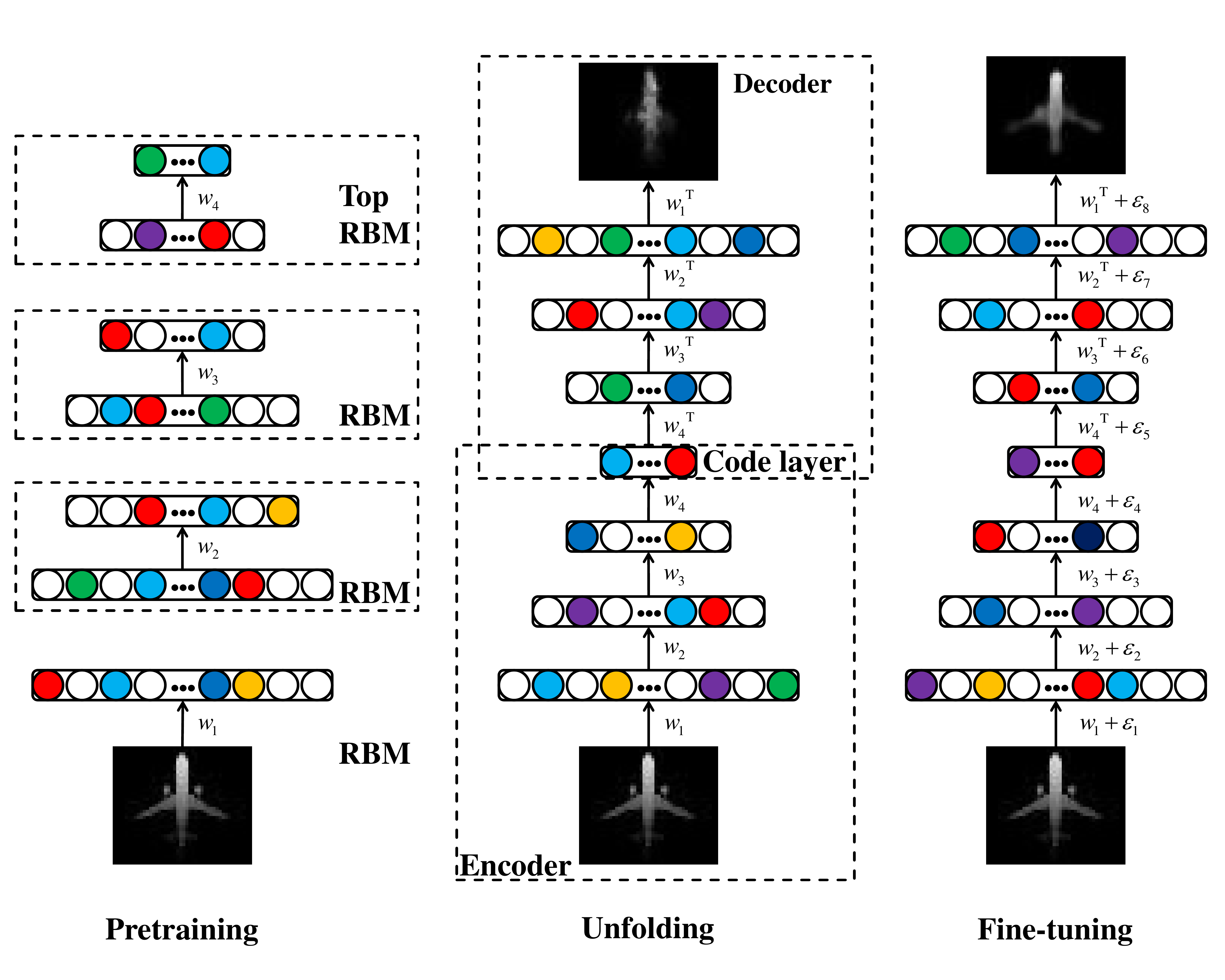}\\
  \caption{Details of autoencoder implemented on depth images. The circles enclosed by rectangle in each layer denote the units with various filling colour indicating different probability that the network assigns to them, and the rectangle's length corresponds to the relative size of dimension on that layer. As we can see, the reconstruction performance becomes much better after doing the fine-tuning procedure compared to the only pretrianing procedure done, which ensures the low-dimensional code layer being a good representation of the 2D image and has a great influence on the retrieval results.}    \label{fig:autoencoder}
\end{figure}

To sum up, the whole autoencoder system is depicted in Fig.~\ref{fig:autoencoder}. Pretraining consists of a stacked RBMs where the hidden units in the previous layer acts as the visible units of the next layer. Then the ``unfolded'' autoencoder initialized by DBN is fine-tuned to obtain a better reconstruction performance. Finally, the code layer that is an efficient representation of the input image is utilized to conduct 3D retrieval.

\subsection{Set-to-Set Distance}
After projecting 3D model and then reconstructing 2D depth images, we get a low-dimensional representation of $S$ with a code set \textbf{C}
\begin{equation}
   \textbf{C}(S)=\lbrace\overrightarrow{C_1} ,\overrightarrow{C_2},\dots,\overrightarrow{C_{Np}}\rbrace,
\end{equation}
where ${Np}$ denotes the number of projection images of each model; and $\overrightarrow{C_i}\ (i=1,2,\dots,Np)$ denotes the coding vector corresponds to the projection $V_i$ with respect to that shape model $S$, defined by
\begin{equation}
   \overrightarrow{C_i}=(c_{i1},c_{i2},\dots,c_{i{Nc}}),
\end{equation}
where $Nc$ denotes the dimensionality of every code vector; $c_{ij}$ is the value of $j$-th dimensionality corresponding to code vector $\overrightarrow{C_i}$.

Based on the effective and efficient autoencoder, we can obtain the quantified distance within each 3D model by defining specific distance method given any two shape model $S_A$ and $S_B$, whose code sets are as follows
\begin{flalign}
\begin{split}
   &\textbf{C}(S_A)=\lbrace\overrightarrow {C_{A_1}},\overrightarrow {C_{A_2}},\dots,\overrightarrow{C_{A_{Np}}}\rbrace\\
   &\textbf{C}(S_B)=\lbrace\overrightarrow {C_{B_1}},\overrightarrow {C_{B_2}},\dots,\overrightarrow{C_{B_{Np}}}\rbrace,\\
\end{split}
\end{flalign}
where $A_i$ and $B_i$ denote the $i$-th projection index of model $S_A$, $S_B$ respectively.

We use one variant of ``Hausdorff Distance'' to define the distance of $S_A$ to $S_B$, given by
\begin{equation}\label{eq:Distance}
   \textit{D}(S_A,S_B)=\frac{1}{Np}\sum_{i=1}^{Np}\min_jd\{\overrightarrow{C_{A_i}},\overrightarrow{C_{B_j}}\},
\end{equation}
where $d\{\overrightarrow{C_{A_i}},\overrightarrow{C_{B_j}}\}$ denotes one specific distance function between two vector, such as p-norm distance in ``Euclidean Space'', algebraic distance, etc.
Depending on the distance of any two models, shape retrieval could be directly done according to the ranked list.

\section{Bag of Features Representation}\label{sec:BoF}

In this Section, we describe the local descriptor formerly implemented by Ohbuchi et al.~\cite{BoF} on 3D shape. Considering that our method autoencoder mentioned above is a global descriptor, it is much reasonable to boost a better performance if combining with a local descriptor. Bag-of-Features (BoF-SIFT) model is selected as the local description for a 3D model. Different from previous work IM-SIFT in~\cite{BoF} that considers the SIFTs~\cite{loweSIFT} of each depth image separately, we put all SIFTs in a single bag, \emph{i.e.}, rotation normalization is not conducted.

We first learn the visual word vocabulary with size of $1500$ in a randomly selected subset of all features via K-means off-line. In order to encode the set of SIFTs in each 3D model, we conduct Vector Quantization proposed in~\cite{SPM} to get a histogram representation that counts the number of SIFTs belonging to each visual word. Before computing the pairwise distance among the models, all the histogram is $L_1$ normalized.
We will display the good property of extraordinary complementarity between autoencoder and BoF-SIFT in Section~\ref{sec:Exp}.

\section{Experiments}\label{sec:Exp}
In this Section, we test our method on two widely used, standard datasets of 3D shapes and compare our results with the-state-of-the-art approaches for 3D shape retrieval. The algorithm is implemented in MATLAB and experiments are carried out on a laptop machine with Intel(R) Core(TM) i5-3210M CPU(2.5GHz) and 4GB memory.

\subsection{Princeton Shape Benchmark(PSB)}\label{sec:PSB}
\begin{figure}[!ht]
   \centering
   \includegraphics[width=0.8\linewidth]{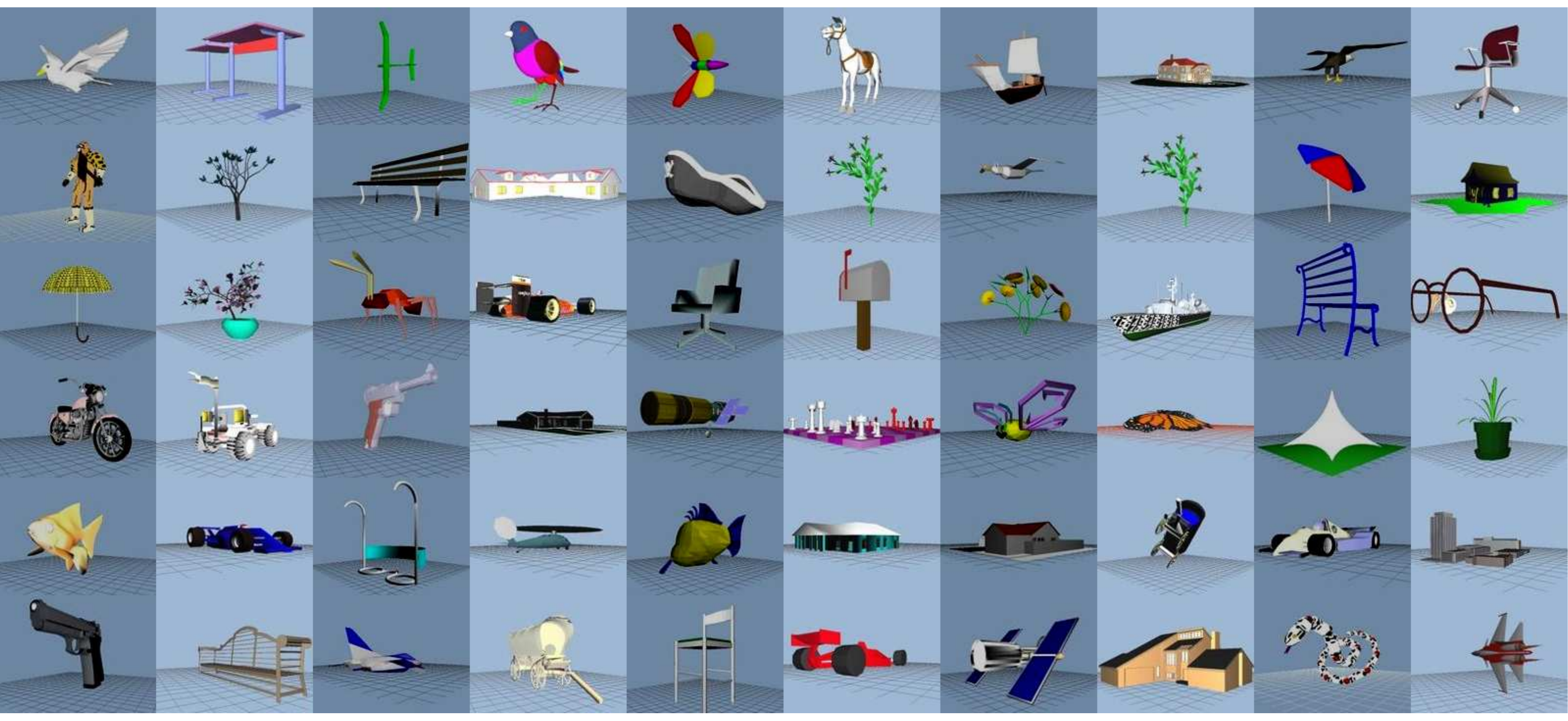}\\
   \caption{Exemplar images randomly chosen from the PSB dataset. The base classification spans a large various of classes including animals, buildings, etc.}\label{fig:PSBData}
\end{figure}

The $Princeton \ Shape\ Benchmark$ \cite{princeton} dataset provides a repository of 3D models and software tools for comparing different shape-based models.
There are totally 1814 models and the base classification is partitioned equally into training and testing sets. The training set with 90 classes, 907 models is used to attain parameters of shape models through training procedure, while the other with 92 classes, equal number of models for comparison with other algorithm. In addition, the number of models belonging to the same class in the base classification varies from each class and ranges from 4 to 50.
Some 3D models from the PSB are randomly selected to be exhibited in Fig.~\ref{fig:PSBData}.

\subsection{Engineering Shape Benchmark(ESB)}
\begin{figure}[!ht]
   \centering
   \includegraphics[width=0.8\linewidth]{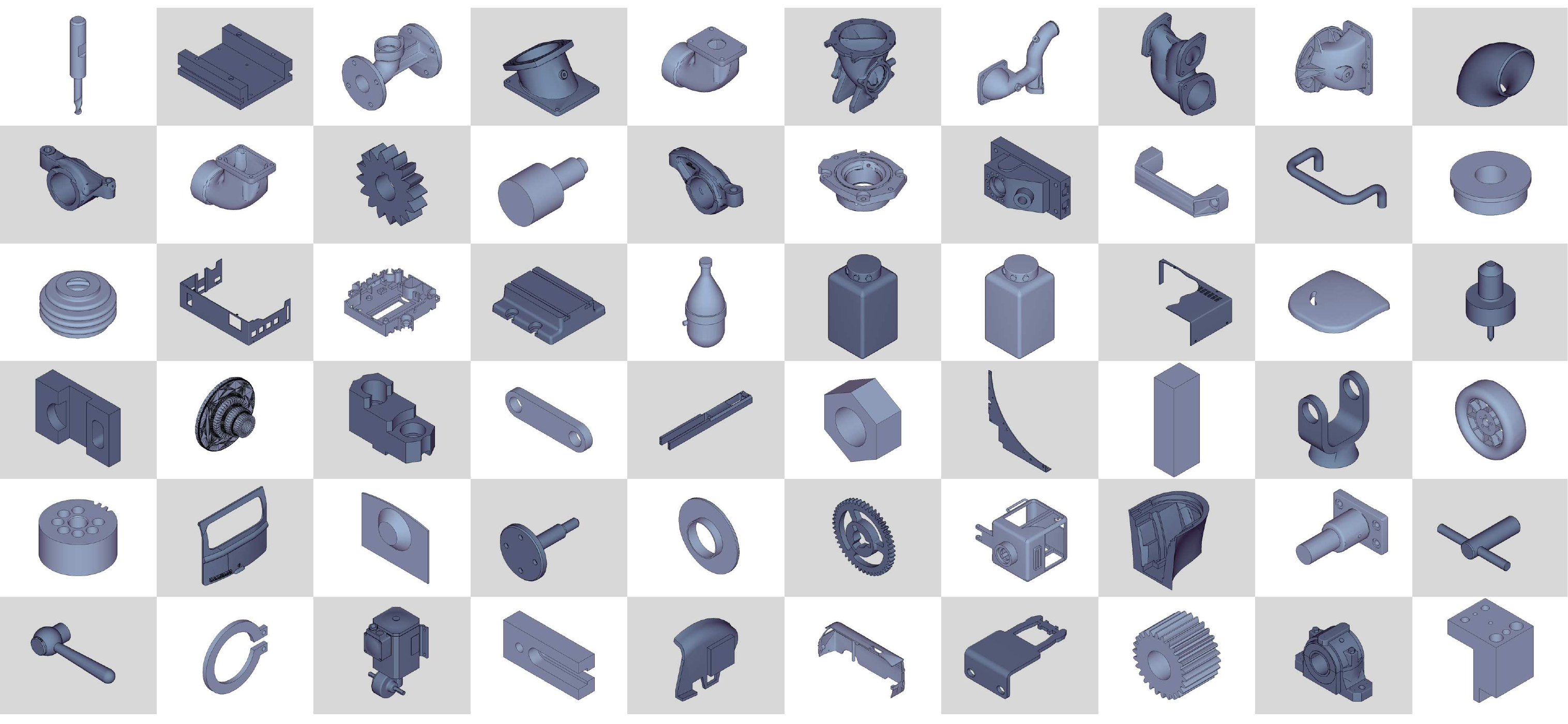}
   \caption{Exemplar images randomly chosen from the ESB dataset. Compared to the PSB dataset, all of the models contained in the ESB are mechanical engineering objects(parts) such as bearing assemblies, spacer, spinner, etc.}\label{fig:ESBData}
\end{figure}

The $Engineering\ Shape\ Benchmark$~\cite{perdue} is particularly proposed to evaluate shape-based searching methods relevant to the mechanical engineering domain.
More specifically, the ESB dataset has totally 867 3D CAD models classified into 45 classes with the number of models ranging from 4 to 58 in a class.
As shown in Fig.~\ref{fig:ESBData}, we randomly select some models in the ESB to display engineering property of the models.

\subsection{Implementation Details}\label{sec:ImDetail}

As described in Section~\ref{sec:DepthPro}, we set the number of each model's projection to 64 $( 8\times 8)$ on the dataset. Then the total raw gray-scale images with real value in the range of $[0,1]$, preprocessed by TILT~\cite{TILT} to eliminate the large orientation variance, served as the visible units of the DBN's first layer.

More specifically, the visible units of the first RBM layer were the normalized value of the depth images' pixels. When training higher level layer, the visible units of a RBM were set to the activation probabilities of the previous RBM's hidden units. As for the hidden units, they had stochastic binary value except the top layer's hidden units, which had stochastic real-valued states calculated from the unit standard deviation Gaussian whose mean value was defined by the input from that RBM's logistic visible units. The real-valued states are in the range $[0,1]$, compared to the binary states either 0 or 1, allowed the low-dimensional codes to take good advantage of continuous data and could avoid unnecessary sampling noise. Note that we trained each RBM for 40 epochs using mini-batches of size 100 and adopted a learning rate of 0.1 for the linear-binary RBMs, 0.001 for the top layer RBM.

With the DBN structure constructed, we initialized an autoencoder with the weights trained from the DBN and fine-tuned them using backpropagation as described in Section~\ref{sec:FineTune}. The autoencoder consisted of an encoder with the designed layers and a symmetric structure for the decoder. The hidden units in the last layer were linear while all the other units were logistic. The deep, well-trained autoencoder was able to find how to convert each depth image into low-dimensional code that leads to a discriminative description and well reconstruction.

Then all the parameters including weights and biases are well-trained in an unsupervised way, we used them to obtain the low-dimensional code for projections of 3D models on the dataset. For the PSB, we constructed an encoder with the layers structure of 5184 $(72 \times 72)$-1000-500-250-30 while a structure of 5184 $(72 \times 72)$-2000-500-100-20 for the ESB. In addition, we only used the testing set to both train the parameters and evaluate our results for the PSB while experiments were done on the whole dataset of the ESB since it provides no training set or testing set.

Finally, we define the distance function as mentioned in Section~\ref{sec:FineTune} as
\begin{equation}
   d\{\overrightarrow{C_{A_i}},\overrightarrow{C_{B_j}}\}={\Arrowvert{\overrightarrow{C_{A_i}}-\overrightarrow{C_{B_j}}}\Arrowvert}_p, \ p=2,
\end{equation}
where ${\Arrowvert{x}\Arrowvert}_p={(|x_1|^p+|x_2|^p+\dots+|x_n|^p)}^\frac{1}{p}$, please note that $x$ is a vector in the $n$-dimensional real vector space $\Re^n$.

\subsection{Evaluation Methods}
The PSB provides open source code for evaluating different algorithm and judging how well the algorithm is compared to others.
When any doubt comes to you, please refer to~\cite{princeton} for more details about definition of every evaluation method.

\textbf{Nearest Neighbor:} the percentage of the closest matches that belong to the same class as the query. This statistic offers us an indication of how well a nearest neighbor classifier could perform. As we can see, higher score represents better performance.

\textbf{First-Tier and Second-Tier:} the percentage of models in the query's class that appear within the top $M$ matches. where $M$ is determined by the size of the query's class.Given that the query's class owns $C$ models, $M=C-1$ for the first-tier and $M=2(C-1)$ for the second tier.



\subsection{Retrieval Results} 
As shown in Table~\ref{tab:GPSBResults} and~\ref{tab:GESBResults}, we compare the global-feature-based autoencoder with the other global descriptors on the two standard datasets to explore the efficacy of using autoencoder to tackle 3D shape retrieval.

\begin{table}[!ht]
\begin{center}
\begin{small}
\caption{Statistic evaluation of global descriptors on Princeton Shape Benchmark}\label{tab:GPSBResults}
\begin{tabular}{|c|c c c|}
  \hline
  Algorithm & NN(\%) & FT(\%) &ST(\%) \\
  \hline
  \hline
  Autoencoder       & $\bm{72.4}$ & $\bm{43.3}$ & $\bm{54.6}$ \\
  GSMD~\cite{GSMD}  &67.1 & 41.8    &52.0 \\
  DESIRE~\cite{Desire}  & 65.8 & 40.4 & 51.3 \\
  LFD~\cite{LFD} & 65.7&38.0&48.7 \\
  \hline
\end{tabular}
\end{small}
\end{center}
\end{table}

\begin{table}[!ht]
\begin{center}
\begin{small}
\caption{Statistic evaluation of global descriptors on Engineering Shape Benchmark}\label{tab:GESBResults}
\begin{tabular}{|c|c c c|}
  \hline
  Algorithm & NN(\%) & FT(\%) &ST(\%) \\
  \hline
  \hline
  Autoencoder       & $\bm{85.7}$ & $\bm{47.9}$ & $\bm{63.1}$ \\
  Hybrid~\cite{Hybrid} & 82.9 & 46.5 & 60.5 \\
  DESIRE\cite{Desire}& 82.3 & 41.7 & 55.0 \\
  LFD~\cite{LFD} & 82.0&40.4&53.9 \\
  \hline
\end{tabular}
\end{small}
\end{center}
\end{table}

We come to a solid conclusion that the autoencoder is more efficient than the other global-features-based methods for 3D shape retrieval.

\subsection{Complementary Property}

Furthermore, based on the knowledge that autoencoder reconstructs global information while BoF-SIFT described in Section~\ref{sec:BoF} captures the local details, a linear combination of them is proposed to boost the retrieval performance. More specifically, we empirically choose the weights as $W_{global}=W_{local}$ for global and local descriptors. 


\begin{table}[!ht]
\begin{center}
\begin{small}
\caption{Statistic evaluation on Princeton Shape Benchmark}\label{tab:PSBResults}
\begin{tabular}{|c|c c c|}
  \hline
  Algorithm & NN(\%) & FT(\%) &ST(\%) \\
  \hline
  \hline
  Autoencoder+BoF-SIFT  & $\bm{77.5}$ & $\bm{52.4}$ & $\bm{65.4}$ \\
  BoF-SIFT~\cite{BoF}               & 71.4 & 45.1 & 57.6 \\
  \hline
  \hline
  CM-BoF+GSMD~\cite{CM-BoF} &75.4 &50.9 &64.0 \\
  PANORAMA~\cite{Panorama} & 75.3 & 47.9 & 60.3 \\
  CM-BoF~\cite{CM-BoF} & 73.1 & 47.0 & 59.8 \\
  \hline
\end{tabular}
\end{small}
\end{center}
\end{table}

\vspace{-0.1in}
We compare our hybrid method (Autoencoder+BoF-SIFT) with the previous state-of-the-art methods including PANORAMA, CM-BoF and CM-BoF+GSMD, which are able to capture both the global and local information of a 3D shape. For the retrieval results displayed in Table \ref{tab:PSBResults} on the PSB dataset, we can find that:
our autoencoder shows pretty well complementary property with the existing local-features-based method BoF-SIFT, whose retrieval results of FT and ST are both improved by more than 7 percent.

\vspace{-0.1in}
\section{Conclusions}\label{sec:CS}
In this paper, we present a novel view-based 3D shape retrieval
method using autoencoder, which is firstly utilized
to 3D shape retrieval. A set of experiments
were carried out to investigate the effectiveness and efficiency of
our method on two standard datasets, which shows that the
autoencoder outperforms other global descriptor on retrieval
results. Furthermore, the experiments demonstrate that the
autoencoder displays good complementarity with the local descriptor,
for linearly combing them achieves the state-of-the-art performance.
Our future work might focus on studying the effect of the proposed
representation with context-based shape similarity method
\cite{Learning Context}.


\vspace{-0.1in}
\section{Acknowledgement}

This work was primarily supported by National Natural
Science Foundation of China (NSFC) (No. 61222308), and Program for New Century Excellent Talents in University (No.NCET-12-0217), Fundamental Research Funds for the Central Universities (No.HUST 2013TS115). Xinggang Wang was supported by Microsoft Research Fellow Award 2012 and Excellent Ph.D. Thesis Founding of HUST 2014.


%
%



%
%
%
\vspace{-0.1in}
\bibliographystyle{splncs}
\bibliography{egbib}

\end{document}